\documentclass{sigchi-ext}
\usepackage[T1]{fontenc}
\usepackage{textcomp}
\usepackage[scaled=.92]{helvet} 
\usepackage{graphicx} 
\usepackage{balance}  
\usepackage{booktabs} 
\usepackage{ccicons}  
\usepackage{ragged2e} 

\def\plaintitle{The Rumour Mill: Making the Spread of Misinformation Explicit and Tangible} 
\def\emptyauthor{}
\def\plainkeywords{Misinformation; NLP; rumour spread; critical design.}

\title{The Rumour Mill: Making the Spread of Misinformation Explicit and Tangible}

\numberofauthors{3}
\author{%
  \alignauthor{%
    \textbf{Nanna Inie}\\
    \affaddr{Lix Technologies} \\
    \affaddr{IT University of Copenhagen} \\
    \affaddr{Copenhagen, Denmark} \\
    \email{nans@itu.dk} } \vfil  \alignauthor{%
    \textbf{Jeanette Falk Olesen}\\
    \affaddr{Aarhus University}\\
    \affaddr{Aarhus, Denmark}\\
    \email{jfo@cc.au.dk} } \vfil \alignauthor{%
    \textbf{Leon Derczynski}\\
    \affaddr{IT University of Copenhagen}\\
    \affaddr{Copenhagen, Denmark}\\
    \email{ld@itu.dk} } }

\definecolor{linkColor}{RGB}{6,125,233}
\hypersetup{%
  pdftitle={\plaintitle},
  pdfauthor={\emptyauthor},
  pdfkeywords={\plainkeywords},
  bookmarksnumbered,
  pdfstartview={FitH},
  colorlinks,
  citecolor=black,
  filecolor=black,
  linkcolor=black,
  urlcolor=linkColor,
  breaklinks=true,
}

\copyrightinfo{\scriptsize Permission to make digital or hard copies of part or all of this work for personal or classroom use is granted without fee provided that copies are not made or distributed for profit or commercial advantage and that copies bear this notice and the full citation on the first page. Copyrights for third-party components of this work must be honored. For all other uses, contact the owner/author(s). \\
{\emph{CHI '20 Extended Abstracts, April 25--30, 2020, Honolulu, HI, USA.}} \\
\copyright~2020 Copyright is held by the author/owner(s). \\
ACM ISBN 978-1-4503-6819-3/20/04. \\
http://dx.doi.org/10.1145/3334480.3383159}

\begin{document}

\maketitle

\RaggedRight{} 

\begin{abstract}
Misinformation spread presents a technological and social threat to society. With the advance of AI-based language models, automatically generated texts have become difficult to identify and easy to create at scale. We present ``The Rumour Mill", a playful art piece, designed as a commentary on the spread of rumours and automatically-generated misinformation. The mill is a tabletop interactive machine, which invites a user to experience the process of creating believable text by interacting with different tangible controls on the mill. The user manipulates visible parameters to adjust the genre and type of an automatically generated text rumour. The Rumour Mill is a physical demonstration of the state of current technology and its ability to generate and manipulate natural language text, and of the act of starting and spreading rumours.
\end{abstract}




\section{Introduction and contribution}
Identifying and verifying misinformation, rumours, and information is difficult, compounded by the emergence of  AI-generated\footnote{Strictly speaking, the recent slew of text generation models are not AI because they are not intelligent. 
Rather, we use this term to situate our methods within the AI research domain.} texts which can be so believable as to be be indistinguishable from human-written text. 
This presents profound social issues, as the spread of misinformation and rumours can have serious consequences. 
With The Rumour Mill we make the process of rumour creation explicit by inviting users to manipulate tangible controls, which alter parameters of an AI-generated rumour. 
In this way, the mill is an opportunity to inform users, as they get to experience for themselves the current possibilities in creating believable rumours automatically.
The prototype makes recent advances in natural language generation visible and accessible to a broad audience, while also making the act of starting and sharing a rumour overt and tangible.

\begin{marginfigure}[0pc]
  \begin{minipage}{\marginparwidth}
    \centering
    \includegraphics[width=1.0\marginparwidth]{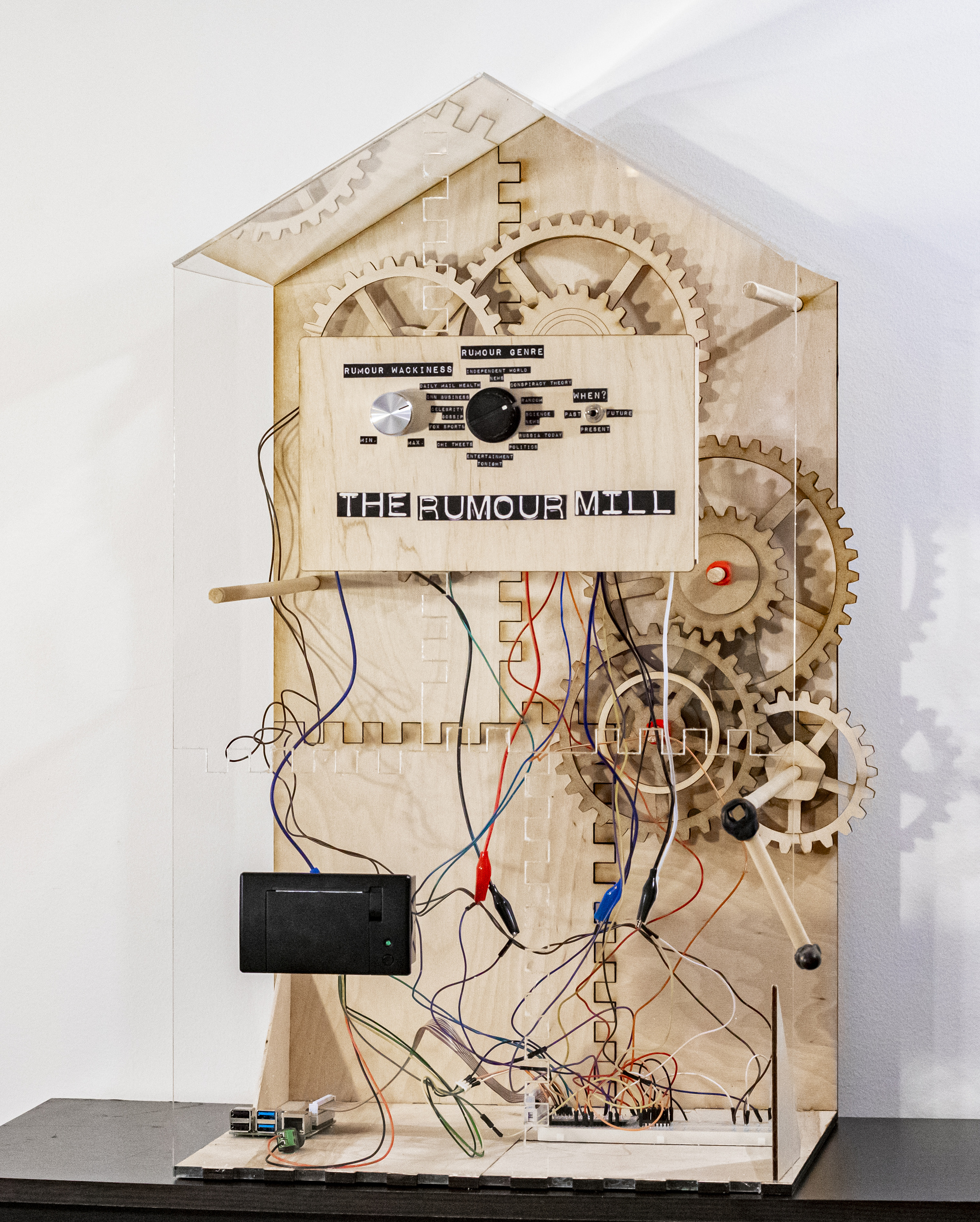}
    \caption{The Rumour Mill makes the process of creating rumours by means of NLP technology explicit, by letting users interact with tangible controls that alter parameters for the generated text.\\ \vspace{1mm} \textbf{Video~demonstration:} \href{https://www.youtube.com/watch?v=L7HBenbJWMw}{youtu.be/L7HBenbJWMw}}~\label{fig:marginfig}
  \end{minipage}
\end{marginfigure}

\section{Background}
The creation and spread of intentionally false and misleading claims undermines social processes of sharing information and collective decision making.
Since society shifted to the internet as its main information source, it has been much easier to become a news publisher.
The subsequent proliferation of information sources has made it easy for inauthentic information to subvert, distract from, and undermine mainstream, authoritative news.
This brings risks. Misinformation has played a role in recent political processes; the 2016 US Election and the Brexit vote both showed signs of interference and misinformation~\cite{howard2016bots,shao2018anatomy}.

People spread misinformation and rumours for a variety of reasons.
Some share false stories in order to gain attention; others in order to blend in with a group sharing stories (particularly in the wake of disasters~\cite{arif2016information,mendoza2010twitter}); and malicious actors will also seed and propagate false narratives, to manipulate opinion or simply to sow dissent and wreak havoc.
However, this is always a covert process.

A further challenge presented by the diversification of information sources is that the origin of a claim becomes less clear, and thus so does its veracity.
To address this, many fact-checking agencies have appeared around the world to organise against and monitor false claims.\footnote{E.g. the IFCN, https://www.poynter.org/ifcn/} 
However this work is Sisyphean; false claims abound, and the cost of creating new ones is very low.
In contrast, the effort need to verify a claim or rumour is high, involving expert skills and having few technological tools to help~\cite{tolmie2017supporting,derczynski2019misinformation}.

Compounding this, automatic tools for generating coherent and believable text have recently become much more proficient~\cite{radford2019language}. Some are rated by humans as better at producing believable, quality text than human authors~\cite{zellers2019defending}. The risk this technology presents has been considered so great that the release of some models was embargoed to give research time to catch up with identifying such text~\cite{radford2019language}. The extra load presented to fact-checkers by this renewable, tireless source is potentially huge.

By making the rumour seeding process publicly visible and explicit, this no longer is a covert process conceived of by the rumour's spreader, but instead a public process; and by making the rumour a tangible object labelled with its source, the spread of the rumour and identification of its nature become explicit. At the same time, people interacting with these rumours should know that the rumours are automatically generated, exposing them to current text generation technology and potentially enhancing their ability to spot automatically-generated or other suspicious claims.

\section{Demo set-up}
\label{sec:setup}

\begin{marginfigure}[0pc]
  \begin{minipage}{\marginparwidth}
    \centering
    \includegraphics[width=1.0\marginparwidth]{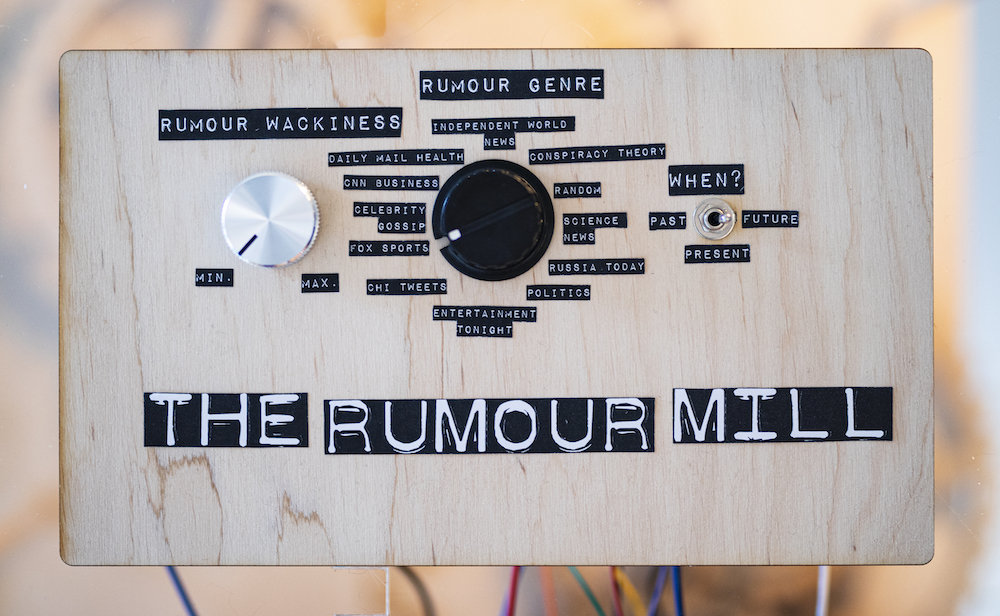}
    \caption{The interaction panel of the mill displaying the input parameters.}~\label{fig:displaypanel}
  \end{minipage}
\end{marginfigure}

\begin{marginfigure}[5pc]
  \begin{minipage}{\marginparwidth}
    \centering
    \includegraphics[width=1.0\marginparwidth]{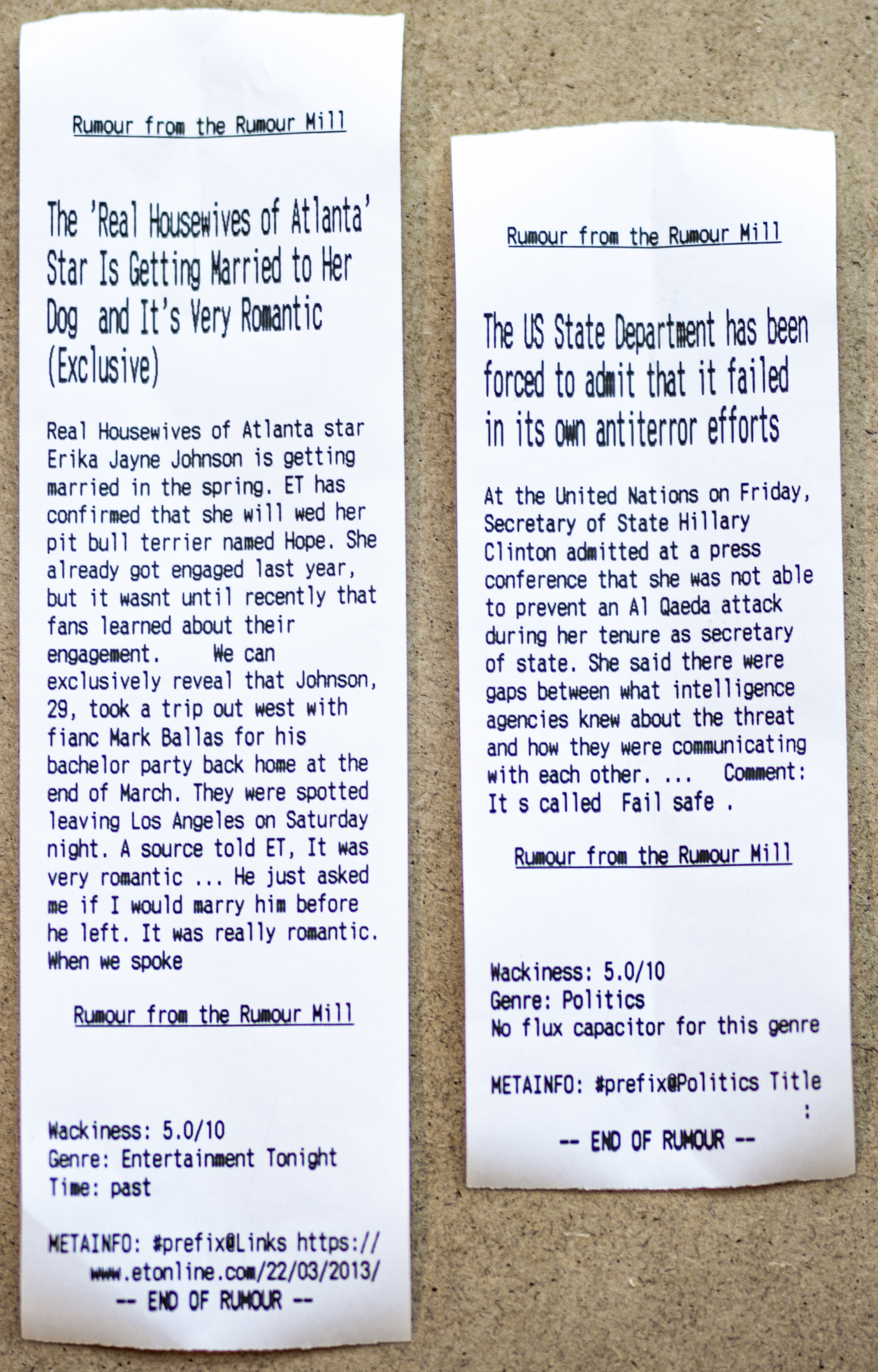}
    \caption{Example rumours printed by the Rumour Mill.}~\label{fig:rumours}
  \end{minipage}
\end{marginfigure}

A wooden and transparent plexiglass box contains a computer and printer and offers a number of input controls. The computer manages rumour generation and delivers rumours via the printer. The input controls allow users to specify or tune the rumour they would like to create, through different modalities (i.e. a potentiometer, a 12-step switch and a toggle switch). The rotary lever is used to initiate rumour generation, explicitly evoking the metaphor of milling, cf. a rumour mill, co-opting the idiom for a ``process in which a group or network of persons originate or promulgate gossip and other unsubstantiated claims."\footnote{From https://en.wiktionary.org/wiki/rumor\_mill} The rumour is then printed on a thermal printer, together with the input settings, and can be thrown away, brought home as a souvenir, or shared on a bulletin board next to the machine.

The mill is made to be large enough that it is awkward to operate covertly. This brings rumour production into an act that is not only tangible for the instigator, but also visible for those nearby. Printed output is marked clearly as being a rumour, both making it easy to source and thus refute, as well as ensuring we are not creating potentially harmful printouts -- an important ethical point, and best-practice in reporting of misinformation~\cite{workshop}.

Power is supplied externally, and rumour generation occurs remotely, as the hardware required to generate this amount of text in real-time is large, noisy, difficult to transport, and warm (therefore not so welcome in the mill itself). However, the mill can store a cache of rumours locally to allow it to operate through temporary network drops.

\section{Technology}

Rumours produced via the Rumour Mill are generated dependent on three user-specified parameters: 1) Rumour wackiness: The user specifies how wacky or conventional the story should be; 2) Rumour genre: The user specifies a genre of the rumour; and 3) "When": The user specifies if the rumour should be from the past, the present, or the future. These buttons are placed on a panel as the main interface of the mill (Figure \ref{fig:displaypanel}).

These inputs are converted to control codes and hyperparameters for a CTRL language model~\cite{keskar2019ctrl}. The ``wackiness" corresponds to the ``temperature" of the language model, where a low-wackiness story will make low-risk word choices that more closely resemble the training data of the CTRL language model, and a high-wackiness story will have more unusual words and thus express less-likely terms, actions, entities and phrases.
The genre is described through control codes. The selected available genres are: politics; conspiracy theory; science news; CNN business; Entertainment Tonight; Daily Mail health; Fox Sports; Independent world news; celebrity gossip; CHI tweets; Russia Today; and a Random setting. The ``When" switch sets the time period from which a story comes and is determined by including a date in CTRL's {\tt Links} control code: contemporary stories come from within the twelve months up to today, past stories from the decade before that period, and future ones from twelve months from tomorrow.  

The rumour text generation itself is a two-step process. When a rumour is requested by ``milling" (turning the lever of the mill) and after a user has manipulated the parameters, firstly a headline is requested from a GPT-2~\cite{radford2019language} instance. 
GPT-2 is a high-quality language model that can be fine-tuned to produce a particular kind of text. 
We tuned it to produce claims and headlines, based on titles from Snopes, Politifact, and Emergent articles~\cite{wang2017liar,ferreira2016emergent}. 
This fine-tuning ensures that headline-style text is generated. 
Secondly, the headline is used as a seed for another language model, CTRL~\cite{keskar2019ctrl}. The CTRL model provides a number of control codes that condition the generated text. 
In this case, the text is tuned to fit the user-input parameters and to continue on from the headline. 
CTRL thus generates a fluent story that matches the requested style, based on the headline claim from GPT-2. 
The output from a milling is one printed rumour, including headline and news blurb.

\section{Conclusion}
Automatically generated text can be used for creating and spreading misinformation and rumours, which can be extremely difficult to identify and verify. The Rumour Mill is a playful object of commentary on the process of artificially generating and manipulating believable rumours. In interacting with The Rumour Mill, the piece provides an overt experience of what is possible with AI-generated texts.

\section{Acknowledgements}
Verif-AI project (Independent Danish Research Foundation); MultiStance project (ITU Copenhagen); industrial researcher fellowship, CIBIS project (Innovation Fund DK).

\balance{} 

\bibliographystyle{SIGCHI-Reference-Format}
\bibliography{sample}

\end{document}